\documentclass[11pt,a4paper]{article}
\pdfoutput=1
\usepackage[hyperref]{naaclhlt2018}
\usepackage{times}
\usepackage{latexsym}

\usepackage{url}

\aclfinalcopy 

\usepackage{amsmath}
\usepackage{multirow}
\usepackage{url}
\usepackage{graphicx}
\usepackage{amssymb}
\usepackage{amsfonts}
\usepackage{todonotes}
\usepackage{fancyvrb}
\usepackage{tikz}
\usepackage{xcolor}
\usepackage{color}
\usepackage{float}

\title{A Novel Embedding Model for Knowledge Base Completion \\ Based on Convolutional Neural Network }

\author{
Dai Quoc Nguyen${}^{1}$, Tu Dinh Nguyen${}^{1}$, Dat Quoc Nguyen${}^{2}$, Dinh Phung${}^{1}$ \\
${}^{1}$ PRaDA Centre, Deakin University, Australia \\
{\tt{{\{dai.nguyen,tu.nguyen,dinh.phung\}@deakin.edu.au}}} \\
${}^{2 }$ The University of Melbourne, Australia \\
 {\tt{{dqnguyen@unimelb.edu.au}}}
}

\begin{document}
\maketitle

\begin{abstract}

In this paper, we propose a novel embedding model, named ConvKB, for  knowledge base completion. Our  model ConvKB advances  state-of-the-art models by employing a convolutional neural network, so that it can capture global relationships and transitional characteristics between entities and relations in  knowledge bases. In ConvKB, each  triple \textit{(head entity, relation, tail entity)} is represented as a 3-column matrix where each column vector represents a triple element. This  3-column matrix is then fed to a convolution layer where multiple filters are operated on the matrix to generate different feature maps. These feature maps are then concatenated into a single feature vector representing the input triple. The feature vector is multiplied with a weight vector via a dot product to return a score. This score is then used to predict whether the triple is valid or not. Experiments show that ConvKB achieves better link prediction performance than previous state-of-the-art embedding models on two benchmark datasets WN18RR and FB15k-237.

\end{abstract}

\section{Introduction}
Large-scale knowledge bases (KBs), such as YAGO \citep{Suchanek:2007}, Freebase \citep{Bollacker:2008} and DBpedia \citep{lehmann2015dbpedia}, are usually databases of triples representing the relationships between entities in the form of fact \textit{(head entity, relation, tail entity)} denoted as \textit{(h, r, t)}, e.g., \textit{(Melbourne, cityOf, Australia)}.    
These KBs are useful resources in many applications such as semantic searching and ranking \citep{kasneci2008naga,Schuhmacher:2014,xiong2017explicit}, question answering \citep{zhang2016question,hao2017end} and machine reading \citep{yang-mitchell:2017}. 
However, the KBs are still incomplete, i.e., missing a lot of valid triples \citep{NIPS2013_5028,West:2014}. 
Therefore, much research work has been devoted towards \textit{knowledge base completion} or \textit{link prediction}  to predict whether a triple \textit{(h, r, t)} is valid or not \citep{bordes2011learning}.

Many embedding models have proposed to learn vector or matrix representations for entities and relations, obtaining state-of-the-art (SOTA) link prediction results \cite{NickelMTG15}. In these embedding models, valid triples obtain lower implausibility scores than invalid triples.
Let us take the well-known embedding model TransE \citep{NIPS2013_5071} as an example. In TransE,  entities and relations are represented by $k$-dimensional vector embeddings.  
TransE employs a transitional characteristic to model relationships between entities, in which it assumes that  if \textit{(h, r, t)} is a valid fact, the embedding of head entity $h$ plus the embedding of relation $r$ should be close to the embedding of tail entity $t$, i.e. $\boldsymbol{v}_h$ + $\boldsymbol{v}_r$ $\approx$ $\boldsymbol{v}_t$
 (here, $\boldsymbol{v}_h$, $\boldsymbol{v}_r$ and $\boldsymbol{v}_t$ are embeddings of $h$, $r$ and $t$ respectively). That is, a TransE score $\|\boldsymbol{v}_h + \boldsymbol{v}_r - \boldsymbol{v}_t\|_p^p$ of the valid  triple \textit{(h, r, t)}  should be close to  $0$ and  smaller than a score $\|\boldsymbol{v}_{h'} + \boldsymbol{v}_{r'} - \boldsymbol{v}_{t'}\|_p^p$ of an invalid  triple \textit{(h', r', t')}. 
The  transitional characteristic in TransE  also implies the global relationships among same dimensional entries of $\boldsymbol{v}_h$, $\boldsymbol{v}_r$ and $\boldsymbol{v}_t$.

Other  transition-based models extend TransE 
to additionally use projection vectors or matrices to translate head and tail embeddings into the relation vector space,  such as: TransH \citep{AAAI148531}, TransR \citep{AAAI159571}, TransD \citep{ji-EtAl:2015:ACL-IJCNLP}, STransE \citep{NguyenNAACL2016}  and TranSparse \citep{JiLH016}. 
Furthermore, DISTMULT \citep{Yang2015} and ComplEx \citep{Trouillon2016} use a tri-linear dot product to compute the score for each triple.  
Recent research has shown that using relation paths between entities in the KBs could help to get contextual information for improving  KB completion performance \citep{lin-EtAl:2015:EMNLP1,luo-EtAl:2015:EMNLP3,guu2015traversing,Toutanova2016,Nguyen2016}.  See other embedding models for KB completion in \citet{Nguyen2017}.

Recently, convolutional neural networks (CNNs), originally designed for computer vision \citep{lecun1998gradient},  have significantly received research attention  in natural language processing \citep{collobert2011natural,kim2014convolutional}.
CNN learns non-linear features to capture complex relationships with a remarkably less number of parameters compared to fully connected neural networks.
Inspired from the success in computer vision, \citet{Dettmers2017} proposed ConvE---the first model applying CNN for the KB completion task.
In ConvE, only $\boldsymbol{v}_h$ and $\boldsymbol{v}_r$ are reshaped and then concatenated into an input matrix which is fed to the convolution layer.
Different filters  of the same $3\times3$ shape are operated over the input matrix to output feature map tensors. These feature map tensors are  then vectorized and mapped into a vector via a linear transformation.
Then this vector is computed with $\boldsymbol{v}_t$ via a dot product to return a score for \textit{(h, r, t)}. See  a formal definition of the ConvE score function in Table \ref{tab:scorefunctions}. 
It is worth noting  that ConvE  focuses on the local relationships among different dimensional entries in each of $\boldsymbol{v}_h$ or $\boldsymbol{v}_r$, i.e., \emph{ConvE does not observe the global relationships among same dimensional entries of an embedding triple ($\boldsymbol{v}_h$, $\boldsymbol{v}_r$, $\boldsymbol{v}_t$)}, so that ConvE ignores the transitional characteristic in  transition-based models, which is one of the most useful intuitions for the task.

In this paper, we present ConvKB---an embedding model which proposes a novel use of CNN for the KB completion task. In ConvKB, each entity or relation is associated with an unique $k$-dimensional embedding. Let $\boldsymbol{v}_h$, $\boldsymbol{v}_r$ and $\boldsymbol{v}_t$ denote  $k$-dimensional embeddings of $h$, $r$ and $t$, respectively. 
For each triple \textit{(h, r, t)}, the corresponding triple of $k$-dimensional  embeddings ($\boldsymbol{v}_h$, $\boldsymbol{v}_r$, $\boldsymbol{v}_t$) is represented as a $k\times3$  input matrix. 
This input matrix is fed to the convolution layer where different filters of the same $1\times3$ shape are used to extract the global relationships among same dimensional entries of the embedding triple.
That is, these filters are repeatedly operated over every row of the input matrix to produce different feature maps. The  feature maps are concatenated into a single feature vector which is then computed with a weight vector via a dot product to produce a score for the triple  \textit{(h, r, t)}. 
This score is used to infer whether the triple \textit{(h, r, t)} is valid or not.

Our contributions in this paper are as follows:
\begin{itemize}

\item We introduce ConvKB---a novel embedding model of entities and relationships for knowledge base completion. 
ConvKB models the relationships among same dimensional entries of the embeddings. This implies that ConvKB generalizes transitional characteristics in  transition-based embedding  models.

\item We evaluate ConvKB  on two benchmark datasets: WN18RR \citep{Dettmers2017} and FB15k-237 \citep{toutanova-chen:2015:CVSC}.  Experimental results show that ConvKB obtains better link prediction performance than previous SOTA embedding models. In particular,   ConvKB obtains the best mean rank  and the highest Hits@10 on WN18RR, and produces the highest mean reciprocal rank and  highest Hits@10 on FB15k-237. 

\end{itemize}

\begin{table}[!t]
\centering
\setlength{\tabcolsep}{0.4em}
\def\arraystretch{1.05}
\resizebox{7.5cm}{!}{
\begin{tabular}{l|l}
\hline
Model & The score function $f(h,r,t)$\\
\hline
\hline
TransE & $\|\boldsymbol{v}_h$ + $\boldsymbol{v}_r$ - $\boldsymbol{v}_t\|_p^p$\\
DISTMULT & $\langle\boldsymbol{v}_h,\boldsymbol{v}_r,\boldsymbol{v}_t\rangle$\\
ComplEx & $\mathsf{Re}\left(\langle\boldsymbol{v}_h,\boldsymbol{v}_r,\overline{\boldsymbol{v}}_t\rangle\right)$\\
ConvE & $g\left(\mathsf{vec}\left(g\left(\mathsf{concat}\left(\widehat{\boldsymbol{v}}_h,\widehat{\boldsymbol{v}}_r\right)\ast\bold{\Omega}\right)\right)\boldsymbol{W}\right)\cdot\boldsymbol{v}_t$\\
\hline
\hline
ConvKB & $\mathsf{concat}\left(g\left([\boldsymbol{v}_h,\boldsymbol{v}_r,\boldsymbol{v}_t]\ast\bold{\Omega}\right)\right)\cdot\bold{w}$\\
\hline
\end{tabular}
}
\caption{The score functions in previous SOTA  models and in our ConvKB model. $\|\boldsymbol{v}\|_p$ denotes the $p$-norm of $\boldsymbol{v}$. $\langle\boldsymbol{v}_h,\boldsymbol{v}_r,\boldsymbol{v}_t\rangle$ = $\sum_i\boldsymbol{v}_{h_i}\boldsymbol{v}_{r_i}\boldsymbol{v}_{t_i}$ denotes a tri-linear dot product. $g$ denotes a non-linear function. $\ast$ denotes a convolution operator. $\cdot$ denotes a dot product. $\mathsf{concat}$ denotes a concatenation operator. $\widehat{\boldsymbol{v}}$ denotes a 2D reshaping of $\boldsymbol{v}$. $\bold{\Omega}$ denotes a set of filters.}
\label{tab:scorefunctions}
\end{table}

\section{Proposed ConvKB model}
\label{sec:model}

A knowledge base $\mathcal{G}$ is a collection of valid factual triples in the form of \textit{(head entity, relation, tail entity)} denoted as $(h, r, t)$ such that $h, t \in \mathcal{E}$ and $r \in \mathcal{R}$ where $\mathcal{E}$ is a set of entities and $\mathcal{R}$ is a set of relations.
Embedding models aim to define a \textit{score function} $f$ giving an implausibility score for each triple $(h, r, t)$ such that valid  triples receive lower scores than invalid triples.
Table \ref{tab:scorefunctions} presents score functions in previous SOTA models.

We denote the dimensionality of embeddings by $k$ such that each embedding triple ($\boldsymbol{v}_h$, $\boldsymbol{v}_r$, $\boldsymbol{v}_t$) are viewed as a matrix $\boldsymbol{A} = [\boldsymbol{v}_h,\boldsymbol{v}_r,\boldsymbol{v}_t] \in \mathbb{R}^{k\times3}$. 
And $\boldsymbol{A}_{i,:} \in \mathbb{R}^{1\times3}$ denotes the $i$-th row of $\boldsymbol{A}$. 
Suppose that we use a filter $\boldsymbol{\omega} \in \mathbb{R}^{1\times3}$ operated on the convolution layer.
$\boldsymbol{\omega}$ is not only aimed  to examine the global relationships between same dimensional entries of the embedding triple ($\boldsymbol{v}_h$, $\boldsymbol{v}_r$, $\boldsymbol{v}_t$), but also to generalize the transitional characteristics in the transition-based models.
$\boldsymbol{\omega}$ is repeatedly operated over every row of $\boldsymbol{A}$ to finally generate a feature map $\boldsymbol{v} = [v_1, v_2, ..., v_k] \in \mathbb{R}^{k}$ as:
\begin{equation}
v_i = g\left(\boldsymbol{\omega} \cdot{\boldsymbol{A}_{i,:}} + b\right) \nonumber
\end{equation}
where 
$b \in \mathbb{R}$ is a bias term and $g$ is some   activation function such as  ReLU. 

\begin{figure}[t]
\centering
\includegraphics[width=0.475\textwidth]{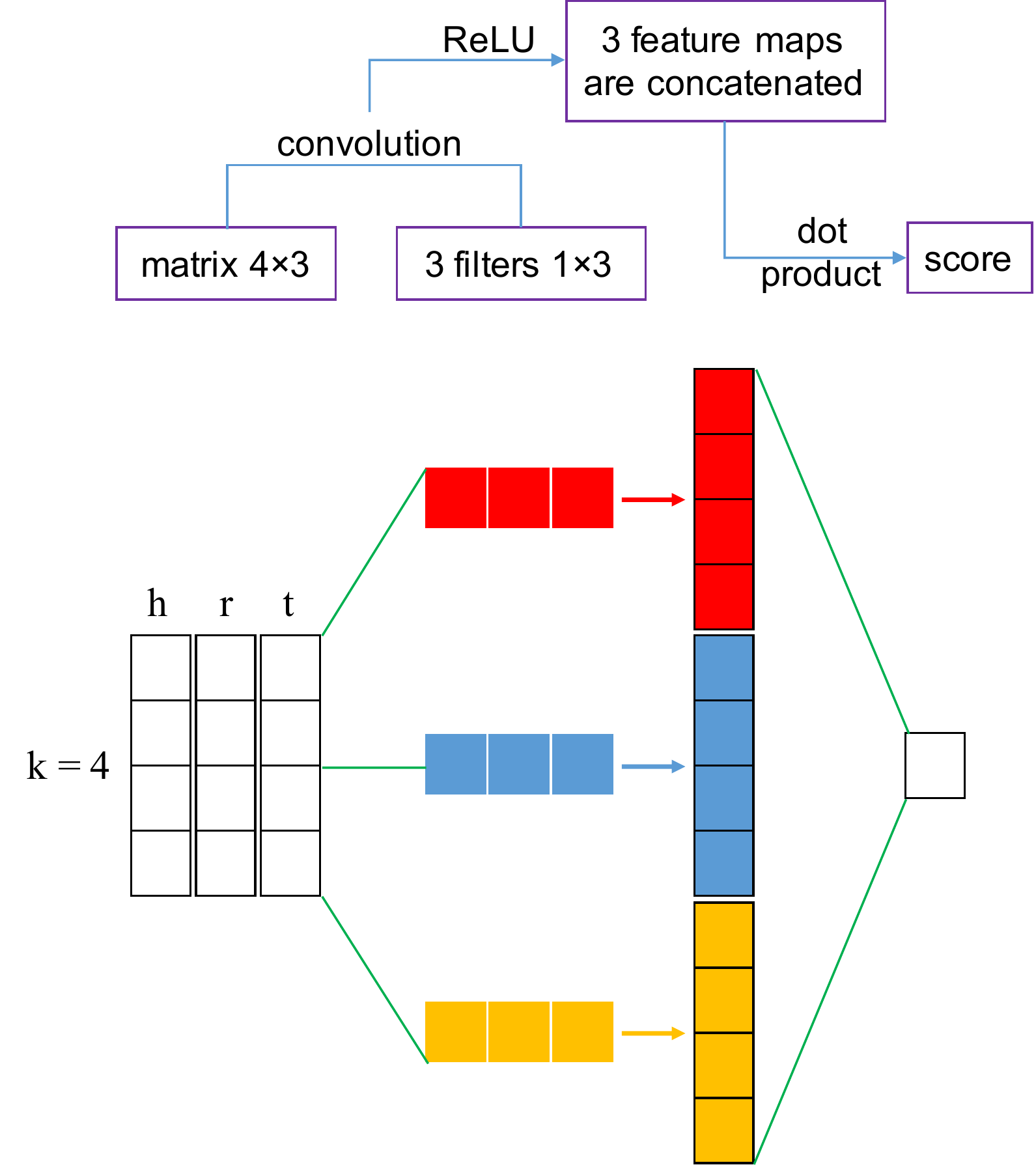}
\caption{Process involved in  ConvKB   (with the embedding size $k=4$,  the number of filters $\tau=3$ and the  activation function $g=$ ReLU for illustration purpose).}
\label{fig:ConvKB}
\end{figure}

Our ConvKB uses different filters $\in \mathbb{R}^{1\times3}$ to generate different feature maps. 
Let $\bold{\Omega}$ and $\tau$ denote the set of filters and the number of filters, respectively, i.e. $\tau = |\bold{\Omega}|$, resulting in  $\tau$ feature maps. 
These $\tau$ feature maps are concatenated into a single vector $\in \mathbb{R}^{\tau k\times1}$ which is then computed with a weight vector $\bold{w} \in \mathbb{R}^{\tau k\times1}$ via a dot product to give a score for the triple $(h, r, t)$. Figure \ref{fig:ConvKB} illustrates the computation process  in  ConvKB.

Formally, we define the ConvKB score function $f$ as follows:
\begin{equation}
f(h,r,t) = \mathsf{concat}\left(g\left([\boldsymbol{v}_h,\boldsymbol{v}_r,\boldsymbol{v}_t]\ast\bold{\Omega}\right)\right)\cdot\bold{w} \nonumber
\end{equation}
where $\bold{\Omega}$ and $\bold{w}$ are shared parameters, independent of $h$, $r$ and $t$; $\ast$ denotes a convolution operator; and $\mathsf{concat}$ denotes a concatenation operator.

If we   only use one filter $\boldsymbol{\omega}$ (i.e. using  $\tau=1$) with a fixed  bias term $b=0$ and the  activation function $g(x)=|x|$ or $g(x)=x^2$, and fix $\boldsymbol{\omega} = [1, 1, -1]$ and $\bold{w} = \textbf{1}$ during training,    ConvKB reduces to the plain TransE model \citep{NIPS2013_5071}. So our  ConvKB  model can be viewed as an extension of   TransE  to further model global relationships. 

We use the Adam optimizer \citep{kingma2014adam} to train ConvKB by minimizing the loss function $\mathcal{L}$ \citep{Trouillon2016} with $L_2$ regularization on the weight vector $\bold{w}$ of the model:
\begin{align}
\mathcal{L} & =  \sum_{\substack{(h,r,t) \in \{\mathcal{G} \cup \mathcal{G}'\}}} \log\left(1 + \exp\left(l_{(h,r,t)} \cdot f\left(h,r,t\right)\right)\right) \nonumber \\
& \ \ \ \ \ \ \ \ \ \ \ \ \ \ \ \ \ \ \ \  \ \ \ \ \ \ \ \ + \frac{\lambda}{2}\|\bold{w}\|^2_2  \nonumber
 \label{equal:objfunc}
  \end{align}
\begin{equation*}
\text{in which, } l_{(h,r,t)} = \left\{ 
  \begin{array}{l}
  1\;\text{for } (h,r,t)\in\mathcal{G}\\
 -1\;\text{for } (h,r,t)\in\mathcal{G}'
  \end{array} \right.
\end{equation*}
here $\mathcal{G}'$ is a collection of invalid triples generated by corrupting valid  triples in $\mathcal{G}$. 

\section{Experiments}

\subsection{Datasets}
We evaluate ConvKB on two benchmark datasets: WN18RR \citep{Dettmers2017} and FB15k-237 \citep{toutanova-chen:2015:CVSC}.
WN18RR and FB15k-237 are correspondingly subsets of two common datasets WN18 and FB15k \citep{NIPS2013_5071}.
As noted by \citet{toutanova-chen:2015:CVSC}, WN18 and FB15k are easy because they contain many reversible relations. So knowing relations are reversible allows us to easily predict the majority of test triples, e.g. state-of-the-art results on both WN18 and FB15k are obtained by using a simple reversal rule as shown in \citet{Dettmers2017}. 
Therefore, WN18RR and FB15k-237 are created to not suffer from this reversible relation problem in WN18 and FB15k, for which the knowledge base completion task is more realistic.  
Table \ref{tab:datasets} presents the statistics of WN18RR and FB15k-237.

\begin{table}[!t]
\centering
\resizebox{7.5cm}{!}{
\setlength{\tabcolsep}{0.4em}
\begin{tabular}{l|lllll}
\hline
\bf Dataset &  $\mid\mathcal{E}\mid$ & $\mid\mathcal{R}\mid$  & \multicolumn{3}{l}{\#Triples in train/valid/test}\\
\hline
WN18RR & 40,943 & 11 & 86,835 & 3,034 & 3,134\\
FB15k-237 & 14,541 & 237 & 272,115 & 17,535 & 20,466\\
\hline
\end{tabular}
}
\caption{Statistics of the experimental datasets.}
\label{tab:datasets}
\end{table}

\begin{table*}[!t]
\centering
\begin{tabular}{l|lll|lll}
\hline
\cline{2-7}
\multirow{2}{*}{\bf Method}& \multicolumn{3}{|c}{\bf WN18RR} & \multicolumn{3}{|c}{\bf FB15k-237}\\
\cline{2-7}
\cline{2-7}
 & MR & MRR & H@10 &   MR & MRR  & H@10 \\
\hline
IRN \citep{ShenHCG17} & -- & -- & -- & \underline{211} & -- & 46.4 \\
KBGAN \citep{Cai2017} & -- & 0.213 & 48.1 & -- & 0.278 & 45.8 \\
DISTMULT \citep{Yang2015} [$\star$] & 5110 & 0.43 & 49 & 254 & 0.241 & 41.9\\
ComplEx \citep{Trouillon2016} [$\star$] & 5261 & \underline{0.44} & \underline{51} & 339 & 0.247 & 42.8\\
ConvE \citep{Dettmers2017} & 5277 & \textbf{0.46} & 48 & 246 & \underline{0.316} & 49.1\\
TransE \citep{NIPS2013_5071} (our results) & \underline{3384} & 0.226 & 50.1 & 347 & 0.294 & 46.5 \\
\hline
Our ConvKB model & \textbf{2554} & 0.248 & \textbf{52.5} & 257 & \textbf{0.396} & \textbf{51.7}\\
\hline
KB$_{LRN}$ \citep{Alberto17} & -- & -- & -- & \textbf{209} & 0.309 & \underline{49.3} \\
R-GCN+ \citep{schlichtkrull2017modeling} & -- & -- & -- & -- & 0.249 & 41.7 \\
Neural LP \citep{YangYC17} & -- & -- & -- & -- & 0.240 & 36.2 \\
Node+LinkFeat \citep{toutanova-chen:2015:CVSC} & -- & -- & -- & -- & 0.293 & 46.2 \\
\hline
\end{tabular}
\caption{Experimental results on WN18RR and FB15k-237 test sets. MRR and H@10 denote the mean reciprocal rank and Hits@10 (in \%), respectively. [$\star$]: Results are taken from \citet {Dettmers2017} where Hits@10 and MRR are rounded to 2 decimal places on WN18RR. The last 4 rows report results of  models that exploit information about relation paths (KB$_{LRN}$, R-GCN+ and Neural LP) or textual mentions derived from a large external corpus (Node+LinkFeat). The best score is in \textbf{bold}, while the second best score is in \underline{underline}.} 
\label{tab:results}
\end{table*}

\subsection{Evaluation protocol}
In the KB completion or link prediction task \citep{NIPS2013_5071}, the purpose is to predict a missing entity given a relation and another entity, i.e, inferring $h$ given $(r, t)$ or inferring $t$ given $(h, r)$.
The results are calculated based on ranking the scores produced by the score function $f$ on test triples.

Following \citet{NIPS2013_5071}, for each valid test triple $(h, r, t)$, we replace either $h$ or $t$ by each of other entities in $\mathcal{E}$ to create a set of corrupted triples.
We use the ``\textbf{Filtered}'' setting protocol \citep{NIPS2013_5071}, i.e., not taking  any corrupted triples that appear in the KB into accounts.
We rank the valid test triple and corrupted triples in ascending order of their scores.
We employ three common evaluation metrics: mean rank (MR), mean reciprocal rank (MRR), and Hits@10 (i.e., the proportion of the valid test triples ranking in top 10 predictions). 
Lower MR, higher MRR or higher Hits@10 indicate better  performance.

\subsection{Training protocol}
We use the common Bernoulli trick \citep{AAAI148531,AAAI159571} to generate the head or tail entities when sampling invalid triples. We also use entity and relation embeddings produced by TransE to \textit{initialize} entity and relation embeddings in ConvKB. We employ a  TransE implementation available at: \url{https://github.com/datquocnguyen/STransE}.  
We train TransE  for 3,000 epochs,
 using a grid search of hyper-parameters: the dimensionality of embeddings $k \in \{50, 100\}$, SGD learning rate $\in \{1e^{-4}, 5e^{-4}, 1e^{-3}, 5e^{-3}\}$, $\mathit{l}_1$-norm or $\mathit{l}_2$-norm, and margin $\gamma \in \{1, 3, 5, 7\}$. 
The highest Hits@10 scores  on the validation set are when using $\mathit{l}_1$-norm, learning rate at $5e^{-4}$, $\gamma$ = 5 and $k$ = 50 for WN18RR, and using $\mathit{l}_1$-norm, learning rate at $5e^{-4}$,  $\gamma$ = 1 and k = 100 for FB15k-237.

To learn our model parameters including entity and relation embeddings, filters $\boldsymbol{\omega}$ and the weight vector $\bold{w}$, we use  Adam \citep{kingma2014adam}  and select its initial learning rate $\in \{5e^{-6}, 1e^{-5}, 5e^{-5}, 1e^{-4}, 5e^{-4}\}$. We use ReLU as the activation function $g$. 
We fix the batch size at 256 and set  the $L_2$-regularizer $\lambda$ at 0.001 in our objective function.
The filters $\boldsymbol{\omega}$ are initialized by a truncated normal distribution or by $[0.1, 0.1, -0.1]$.
We select the number of filters $\tau \in \{50, 100, 200, 400, 500\}$.
We run ConvKB up to 200 epochs and use  outputs from the last epoch for evaluation. 
The highest Hits@10 scores on the validation set are obtained when using $k$ = 50, $\tau = 500$, the truncated normal distribution for filter initialization, 
and the initial learning rate at $1e^{-4}$ on WN18RR; and k = 100, $\tau = 50$, $[0.1, 0.1, -0.1]$ for filter initialization, and the initial learning rate at $5e^{-6}$ on FB15k-237.

\subsection{Main experimental results}
Table \ref{tab:results} compares the experimental results of our ConvKB model with previous published results, using the same experimental setup. Table \ref{tab:results} shows that ConvKB obtains the best MR and highest Hits@10 scores on WN18RR and also the highest MRR and Hits@10 scores on FB15k-237. 

ConvKB does better than the closely related model TransE on both experimental datasets, especially on  FB15k-237 where ConvKB gains significant improvements of $347-257 = 90$ in MR (which is  about 26\% relative improvement) and $0.396 - 0.294 = 0.102$ in MRR (which is   34+\% relative improvement),  and also obtains $51.7 - 46.5 = 5.2$\% absolute   improvement in Hits@10. 
Previous work shows that TransE obtains very competitive results \cite{lin-EtAl:2015:EMNLP1,nickel2016holographic,Trouillon2016,Nguyen2016}. 
However, when comparing  the CNN-based embedding model ConvE  with other models, 
\citet{Dettmers2017} did not experiment with TransE.   
We reconfirm previous findings that TransE in fact is a strong baseline model, e.g., TransE obtains better MR and Hits@10 than ConvE on WN18RR.

ConvKB   obtains  better scores than ConvE on both  datasets (except  MRR  on WN18RR and MR on FB15k-237), thus showing the usefulness of taking transitional characteristics into accounts.  In particular, on   FB15k-237,  ConvKB achieves improvements of $0.394-0.316 = 0.078$ in MRR (which is about 25\% relative improvement) and $51.7 - 49.1 = 2.6$\% in Hits@10, while both ConvKB and ConvE produce similar MR scores.  ConvKB also obtains  25\% relatively higher MRR score than the relation path-based model KB$_{LRN}$ on FB15k-237. In addition, ConvKB  gives better  Hits@10 than KB$_{LRN}$, however, KB$_{LRN}$ gives better MR than ConvKB. We  plan to   extend ConvKB with relation path information to obtain better  link prediction performance  in future work.

\section{Conclusion}
In this paper, we propose a novel embedding model ConvKB for the knowledge base completion task.
ConvKB applies the convolutional neural network to explore the global relationships among same dimensional entries of the entity and relation embeddings, so that ConvKB generalizes the transitional characteristics in the transition-based embedding models.
Experimental results show that our  model ConvKB outperforms other state-of-the-art models on two benchmark datasets WN18RR and FB15k-237.  Our code is available at: \url{https://github.com/daiquocnguyen/ConvKB}.

We also plan to extend ConvKB for a new application where we could formulate data in the form of triples. For example, inspired from the work by \citet{Vu-etal-ECIR2017} for search  personalization,  we can also  apply ConvKB to model  \textit{user}-oriented relationships between submitted 
\textit{queries} and \textit{documents} returned by  search engines, i.e. modeling triple representations (query, user, document). 

\section*{Acknowledgments}
This research was partially supported by the Australian Research Council (ARC) Discovery Grant Project DP160103934.

\bibliographystyle{acl_natbib}
\bibliography{references}

\end{document}